\documentclass[pdflatex,sn-mathphys-num]{sn-jnl}

\usepackage{graphicx}
\usepackage{multirow}
\usepackage{amsmath,amssymb,amsfonts}
\usepackage{mathrsfs}
\usepackage[title]{appendix}
\usepackage{xcolor}
\usepackage{textcomp}
\usepackage{manyfoot}
\usepackage{booktabs}
\usepackage{algorithm}
\usepackage{algorithmicx}
\usepackage{algpseudocode}
\usepackage{listings}
\usepackage{url}
\usepackage{tabularx}
\usepackage{bbding} 
\usepackage{pifont} 
\usepackage{tikz}

\begin{document}

	\title[Neurodevelopmental assessment in preterm infants]{Deep learning-based neurodevelopmental assessment in preterm infants}
	
	\author[1]{\fnm{Lexin} \sur{Ren}}\email{251611018@sust.edu.cn}
	\equalcont{These authors contributed equally to this work.}

	\author[1]{\fnm{Jiamiao} \sur{Lu}}\email{241612058@sust.edu.cn}
    \equalcont{These authors contributed equally to this work.}
	
	\author*[1]{\fnm{Weichuan} \sur{Zhang}}\email{zwc2003@163.com}
	
	\author*[2]{\fnm{Benqing} \sur{Wu}}\email{wubenqing783@126.com}
	
	\author*[3]{\fnm{Tuo} \sur{Wang}}\email{wt1972@tom.com}
	
	\author*[1]{\fnm{Yi} \sur{Liao}}\email{yi.liao@griffith.edu.au}
	
	\author*[4]{\fnm{Jiapan} \sur{Guo}}\email{j.guo@umcg.nl}
	
	\author*[5]{\fnm{Changming} \sur{Sun}}\email{changming.sun@csiro.au}

	\author*[2]{\fnm{Liang} \sur{Guo}}\email{419221185@qq.com}

	\affil[1]{\orgdiv{Image Computing Laboratory}, \orgname{Shaanxi University of Science and Technology}, \city{Xi'an}, \state{Shaanxi Province}, \country{China}}
	
    \affil[2]{\orgdiv{Department of Neonatology}, \orgname{Shenzhen University of Advanced Technology General Hospital}, \city{Shenzhen}, \country{China}}
	
	\affil[3]{\orgdiv{Department of Neurosurgery}, \orgname{The First Affiliated Hospital of Xi'an Jiaotong University}, \city{Xi'an}, \country{China}}
	
    \affil[4]{\orgdiv{Department of Radiotherapy}, \orgname{University of Groningen, University Medical Center Groningen}, \orgaddress{\street{Hanzeplein 1}, \postcode{9713 GZ}, \city{Groningen}, \country{The Netherlands}}}
	
	\affil[5]{\orgname{CSIRO Data61}, \orgaddress{\street{PO Box 76}, \city{Epping}, \state{NSW 1710}, \country{Australia}}}
	
	\abstract{
		Preterm infants (born between 28 and 37 weeks of gestation) face elevated risks of neurodevelopmental delays, making early identification crucial for timely intervention. While deep learning-based volumetric segmentation of brain MRI scans offers a promising avenue for assessing neonatal neurodevelopment, achieving accurate segmentation of white matter (WM) and gray matter (GM) in preterm infants remains challenging due to their comparable signal intensities (isointense appearance) on MRI during early brain development. To address this, we propose a novel segmentation neural network, named Hierarchical Dense Attention Network. Our architecture incorporates a 3D spatial-channel attention mechanism combined with an attention-guided dense upsampling strategy to enhance feature discrimination in low-contrast volumetric data. Quantitative experiments demonstrate that our method achieves superior segmentation performance compared to state-of-the-art baselines, effectively tackling the challenge of isointense tissue differentiation. Furthermore, application of our algorithm confirms that WM and GM volumes in preterm infants are significantly lower than those in term infants, providing additional imaging evidence of the neurodevelopmental delays associated with preterm birth. The code is available at: \url{https://github.com/ICL-SUST/HDAN}.
	}
	
	
	\maketitle
	
	\section{Introduction}
	Preterm infants, defined as neonates born between 28 and 37 weeks of gestation, account for approximately $11\%$ of all deliveries worldwide~\cite{Ohuma2023}.
	Gestational age at birth is strongly correlated with neurological outcomes, and prematurity is recognized as a major risk factor for neurodevelopmental delay~\cite{2005neurologic,2015OutcomesPrematrueInfant,Mitha2024,Nivins2025}.
	Infants born at lower gestational ages are particularly susceptible to cognitive and language impairments, as well as motor dysfunction~\cite{2022MouseModel,Loeffler2025}.
	Given that current treatments for neurodevelopmental disorders are largely ameliorative rather than curative, the early identification of neuro-markers associated with neurodevelopmental delay is of critical importance~\cite{Zhao2023,Vanderhasselt2024}.
	Such early detection would enable the development of targeted, preemptive intervention strategies aimed at improving long-term outcomes or potentially preventing these disorders altogether.
	
	Infant brain MRI segmentation, which delineates brain tissues into mutually exclusive regions, is a fundamental step in the quantitative analysis of early brain development~\cite{2019skipconnect,Bethlehem2022}.
	By computing the volumetric measurements of different tissue types, researchers can quantitatively evaluate developmental trajectories during infancy~\cite{Rossetti2025}.
	Traditionally, brain segmentation has been performed manually by experienced clinicians~\cite{Jain2024NeuroSci}.
	However, this approach is labor-intensive, time-consuming, and susceptible to both intra- and inter-observer variability~\cite{2018VoxResNet,2018review,DeRosa2024,Wasserthal2025}.
	The reliance on highly skilled experts and substantial manual effort has hindered large-scale studies and slowed progress in early neurodevelopmental assessment.

	With the recent integration of artificial intelligence (AI) techniques~\cite{jing2021novel, lu2022image, bao2022corner, tang2022joint, wang2024unbiased, lei2024semi, liao2022asrsnet,jing2022recent, pan2024pseudo,jing2022image,zhang2024re,6507646,shui2012noise, zhang2017noise, zhang2015contour, zhang2020corner, zhang2019discrete, zhang2014corner, zhang2019corner, li2023traffic, qiu2021recurrent, liu2023efficient, jing2023ecfrnet, wang2020corner, wang2018survey, li2019multi, liu2024aekan, ren2024few, zhang2021ndpnet, li2023mutual, gao2020fast,mital2023neural, zhang2023image, duan2023learned,islam2023background, liao2025learning, wang2025principal,zheng2023fully,lu2023track,liao2025dynamic,ren4962361adaptive,an2023edge,xie2026second,song2025efficient,wang2025feature} into medical image analysis, automated MRI segmentation methods based on neural networks have emerged as a promising solution to overcome the inherent limitations of manual segmentation~\cite{2016autoSeg,2015FCN,Richter2022,Kim2024}.
	Recent studies~\cite{2015UNet,20183DSeg,Kim2024,Wasserthal2025} have demonstrated that deep learning-based approaches can achieve comparable or even superior accuracy to expert annotations while significantly reducing analysis time, thereby facilitating large-scale population studies and enabling more efficient early identification of neurodevelopmental risk markers in preterm infants.
	
	\begin{figure}[h]
		\centering
		\includegraphics[width=0.9\columnwidth]{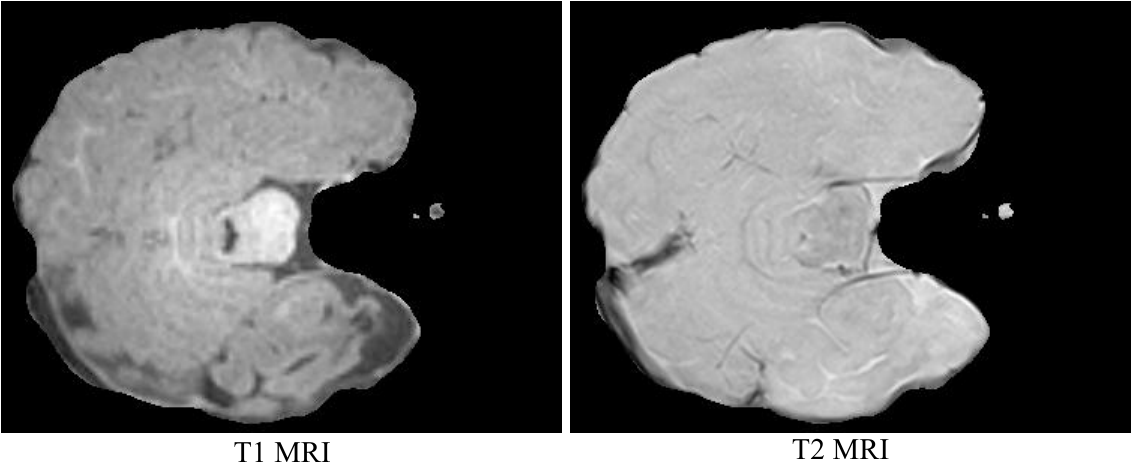}
		\caption{MRI data of an preterm infant subject scanned at $9$ months old (isointense phase).
			From left to right: T1 MRI and T2 MRI. It observed that the intensity level between white matter and gray matter is extremely similar, resulting in the low tissue contrast and thus making the segmentation a difficult challenge.}
		\label{fig1}
	\end{figure}
	
	However, accurately segmenting white matter and gray matter in MRI scans of preterm infants remains a significant challenge.
	The first year of life is the most dynamic phase of postnatal human brain development, during which brain tissue grows rapidly, and cognitive as well as motor functions undergo substantial maturation~\cite{2008MRIstudyl,Mhlanga2023}.
	At this stage, white matter and gray matter display nearly identical intensity levels on both T1- and T2-weighted MRI images, leading to extremely low tissue contrast and making accurate segmentation particularly difficult~\cite{2009autosegNewborn,Mhlanga2023}.
	Therefore, this stage is also referred to as the isointense phase~\cite{Barkovich1988,Dubois2014,Makropoulos2018}.
	For term infants, the isointense phase typically emerges at approximately 6 to 8 months after birth~\cite{Mhlanga2023FrontMed,Baljer2024HumBrainMapp};
	however, for preterm infants with delayed brain development, the timing of this phase remains uncertain, which undoubtedly further complicates the segmentation process.
	Fig.~\ref{fig1} shows examples of T1-weighted and T2-weighted MRI images acquired from a preterm infant at approximately 9 months of age.
	It can be observed that white matter and gray matter exhibit similar intensity levels, particularly in cortical regions, resulting in minimal tissue contrast and making tissue segmentation particularly challenging.
	
	To the best of our knowledge, there are currently two main types of convolutional neural networks (CNNs) developed for volumetric infant brain segmentation.
	The first category consists of modified variants of 2D CNNs~\cite{2013deepfeaturelearning,2015SemanticSeg,2015deepCNN,Wu2025}, which process individual slices or orthogonal planes (i.e., axial, coronal, and sagittal) to approximate three-dimensional spatial information.
	
	Although these approaches have shown promising preliminary results, they cannot fully exploit the inherent 3D contextual information, thereby limiting their ability to achieve highly accurate object segmentation.
	The second category employs true 3D CNNs to segment objects directly from volumetric data, demonstrating compelling performance~\cite{20163DUNet,2016VNet,Chen2023,Kim2024}.
	However, efficiently training these deep 3D networks remains a challenge due to the high dimensionality of the data and the limited number of available training samples.
	
	To bridge a critical gap in the early diagnosis of neurodevelopmental risks on brain MRI images among preterm infants, this paper introduces an automated 3D CNN-based model named HDAN for infant brain MRI volumetric segmentation.
	To ensure robust feature extraction under limited training data, the proposed network adopts a standard Convolution-Batch Normalization-Activation design sequence to maintain training stability.
	Furthermore, by integrating the feature extractor with advanced 3D skip-connections, the network enhances feature representation capability, making it specifically suitable for preterm infant brain MRI images with low tissue contrast.
	In addition, we propose a 3D attention module to enlarge the intensity-level gap between white matter and gray matter, thereby improving the model’s recognition capability on low-contrast images.
	
	The main contributions in this work can be summarized as follows:
	\begin{itemize}
		\item A novel automated 3D CNN-based segmentation model, HDAN, is proposed to address the challenging segmentation of low-contrast white matter and gray matter in MRI images.
		\item An Attention-Guided Dense Upsampling Block (AGDUB) is introduced to effectively recover fine-grained spatial details during the volumetric reconstruction phase, mitigating information loss typically caused by repeated down-sampling operations.
		\item To improve the feature representation capability, the proposed method integrates the inner feature extractor with 3D skip connections, making the model specifically suitable for feature extraction from preterm infant brain MRI images with low tissue contrast.
		\item To enhance the model's recognition capability on low-contrast images, a 3D attention module is proposed to enlarge the intensity-level gap between gray and white matter on preterm infant brain MRI scans.
		\item Quantitative experiments on the benchmark dataset demonstrate that the proposed model not only obtains superior segmentation performance but also effectively delineates tissue boundaries in low-contrast infant brain MRI scans.
	\end{itemize}
	
	\section{Related Work}
	\subsection{3D Features for Volumetric Segmentation}
	Early brain MRI segmentation methods~\cite{2013deepfeaturelearning,2015SemanticSeg,2015deepCNN} primarily used 2D image processing.
	This approach limited their capacity to capture volumetric contextual information~\cite{Avbersek2022}. Utilizing inter-slice correlations improves the accuracy of 3D volumetric segmentation~\cite{ZhangY2022}.
	Both classical and recent studies~\cite{2015learning,Zhou2023nnFormer,Xia2025Review} have shown that modeling 3D features or context significantly improves segmentation performance compared to traditional 2D slice-based methods.
	Based on these insights, many 3D convolutional architectures have been developed for biomedical volumetric segmentation.
	These include early lesion detection methods~\cite{2016autoDetec}, encoder–decoder models such as 3D U-Net~\cite{20163DUNet} and V-Net~\cite{2016VNet}, as well as large-cohort multi-scale frameworks~\cite{2017efficient, 20183DFCN, Billot2023}.
	For example, the 3D U-Net~\cite{20163DUNet} employs skip connections to enhance up-sampling.
	Dual-path CNNs~\cite{2017efficient} have also been developed for accurate brain lesion segmentation~\cite{Fang2022}.
	Although these models demonstrate clear advantages of 3D over 2D feature learning in volumetric brain image segmentation, most adopt the conventional 3D convolutional designs.
	However, optimizing these deep networks on limited medical datasets remains a challenging task that requires efficient architectural strategies.
	
	\subsection{3D-CNN for Volumetric Segmentation}
	Numerous studies~\cite{2017HoughCNN,2016deepMRI, 2015learning,20163DUNet, 2016VNet, 2019skipconnect, 20183DSeg} have explored the use of 3D CNNs for biomedical volumetric segmentation~\cite{Hatamizadeh2022UNETR,Xia2025Review,Zhou2023}.
	Hough-CNN~\cite{2017HoughCNN} integrates a 3D CNN with a Hough voting mechanism for 3D segmentation.
	However, this method is not end-to-end and is mainly restricted to segmenting compact, blob-like structures~\cite{Zhou2023nnFormer,Xia2025Review}.
	Kleesiek et al.~\cite{2016deepMRI} proposed an end-to-end 3D CNN for volumetric segmentation, but the network is shallow and cannot capture multiscale structures~\cite{Isensee2021,Niyas2022}.
	To address this limitation, Moeskops et al.~\cite{2016autoSeg} introduced a multiscale CNN for infant brain tissue segmentation.
	Chen et al.~\cite{2018VoxResNet} introduced residual learning to improve the training of fully convolutional networks (FCNs) for adult brain tissue segmentation.
	However, this approach ignores the information loss caused by pooling operations in FCNs, which can reduce segmentation accuracy.
	To overcome this, 3D-FCN~\cite{20183DSeg} was proposed to enhance segmentation performance.
	However, it still struggles with small structures and precise localization due to coarse feature maps and resolution loss from repeated pooling~\cite{Lilhore2025}.
	These limitations motivate our design of an adaptive 3D attention mechanism that dynamically reweights volumetric features across both spatial and channel dimensions.
	
	Although the aforementioned methods increase network capacity, they also present a common challenge: training becomes more difficult as network depth and complexity increase~\cite{Xia2025Review}.
	To address this, 3D-SkipDenseSeg~\cite{2019skipconnect} incorporates skip connections~\cite{He2016} within the fully convolutional DenseNet~\cite{Huang2017} architecture, allowing information to be concatenated from lower to higher dense blocks.
	These skip connections enable the network to leverage multi-level contextual information, thereby improving segmentation accuracy.
	However, stacked deconvolution layers result in a large number of learnable parameters, leading to increased memory consumption and training time~\cite{Lilhore2025}.
	
	More recently, Transformer-based frameworks have advanced medical image segmentation by enabling global contextual modeling beyond the local receptive fields of convolutional networks~\cite{Zhou2023}.
	The self-configuring nnU-Net~\cite{Isensee2021} established a strong 3D segmentation baseline, while UNETR~\cite{Hatamizadeh2022} and Swin UNETR~\cite{Hatamizadeh2022SwinUNETR} introduced hierarchical attention mechanisms to effectively capture long-range dependencies in volumetric data.
	These developments complement convolutional designs and have motivated research into hybrid CNN-Transformer architectures.
	
	\begin{figure*}[h]
		\centering
		\includegraphics[width=\textwidth]{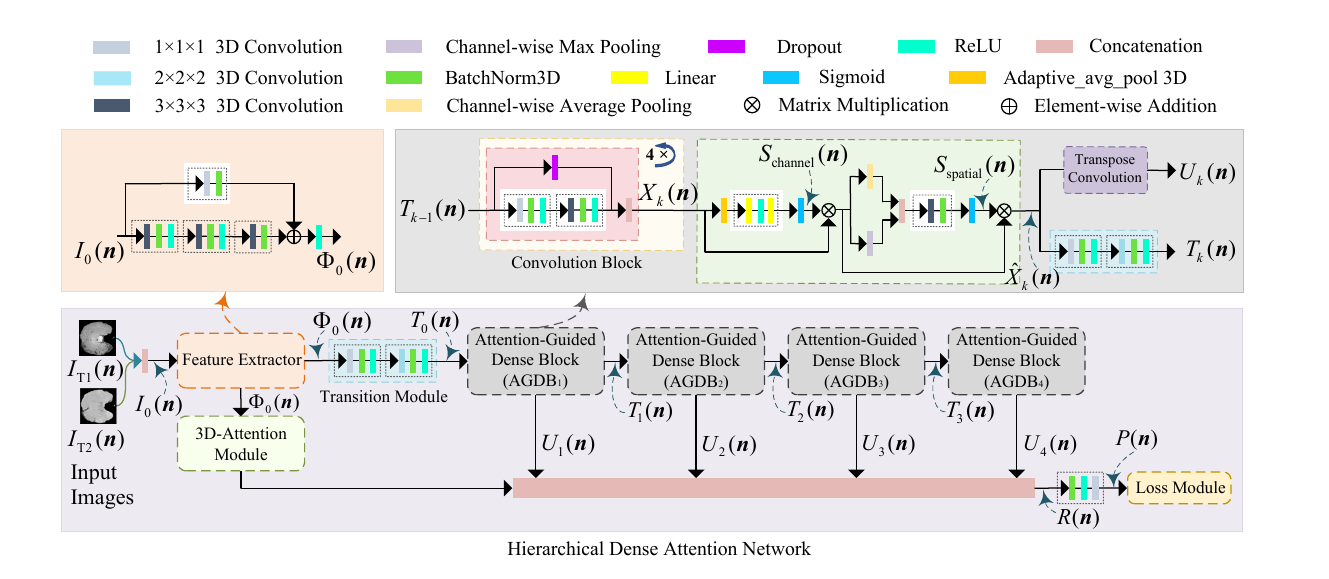}
		\caption{
			The proposed architecture of \textbf{HDAN} (Hierarchical Dense Attention Network) for preterm infant brain volumetric segmentation.
			The framework consists of three main parts: 
			\textbf{(a) Overall Architecture:} The multi-modal inputs are fused into $I_0(\boldsymbol{n})$ and processed by a hierarchical encoder-decoder structure.
			Multi-scale features are recursively extracted via four Attention-Guided Dense Blocks (AGDBs) and aggregated for dense prediction.
			\textbf{(b) Feature Extractor:} Detailed structure of the initial residual feature extraction module generating $\Phi_0(\boldsymbol{n})$.
			\textbf{(c) AGDB Internal Details:} Illustration of the internal information flow, highlighting the parallel generation of the next-stage feature $T_k(\boldsymbol{n})$ via the \textbf{Transition Module} and the upsampled feature $U_k(\boldsymbol{n})$ for global fusion.
		}
		\label{fig2}
	\end{figure*}
	
	\section{The Proposed Method}
	
	\subsection{Problem Definition}
	Given multi-modal MRI scans of infant brains, this study aims to perform voxel-wise segmentation that classifies each voxel into distinct tissue categories, including white matter (WM), gray matter (GM), and cerebrospinal fluid (CSF).
	Formally, let the input MRI volume be denoted as $I = \{I_{T1}, I_{T2}\} \in \mathbb{R}^{2\times D\times H\times W}$, where the two channels correspond to the co-registered T1- and T2-weighted modalities.
	The segmentation network learns a mapping function $\mathcal{F}: I \rightarrow P$, where $P \in \mathbb{R}^{C\times D\times H\times W}$ is the predicted probability map with $C$ tissue classes.
	The task is optimized by minimizing the cross-entropy loss between the predicted distribution and the ground truth labels.
	
	\subsection{Overall Framework}
	As illustrated in Fig.~\ref{fig2}, the proposed Hierarchical Dense Attention Network (HDAN) adopts a recursive encoder-decoder strategy.
	Unlike conventional networks, HDAN introduces a parallel branching strategy within each encoding stage to simultaneously propagate deep semantic context ($T_k$) and fine-grained structural details ($U_k$).
	Let the spatial coordinate of a voxel be denoted as $\boldsymbol{n} = (x, y, z)$.
	The T1- and T2-weighted MRI volumes are channel-wise concatenated to form the unified input $I_0(\boldsymbol{n})$.
	
	\subsubsection{Encoding Path}
	The encoding process begins with a Feature Extractor $\mathcal{F}(\cdot)$, which maps the input $I_0(\boldsymbol{n})$ to the initial high-resolution feature maps $\Phi_0(\boldsymbol{n})$.
	To initiate the hierarchical representation, $\Phi_0(\boldsymbol{n})$ passes through an initial Transition Module $\mathcal{T}_{\mathrm{init}}(\cdot)$ to produce the base feature $T_0(\boldsymbol{n})$.
	Subsequently, the network recursively processes the features through four Attention-Guided Dense Blocks (AGDBs).
	Formally, for the $k$-th stage ($k=1, \dots, 4$), each AGDB receives the transition feature $T_{k-1}(\boldsymbol{n})$ from the previous stage.
	Inside each AGDB, the feature undergoes dense convolution and 3D attention refinement to generate an intermediate refined feature $\hat{X}_k(\boldsymbol{n})$.
	Crucially, as shown in Fig.~\ref{fig2}, this refined feature is split into two parallel branches:
	\begin{itemize}
		\item Hierarchical Transition Branch: $\hat{X}_k(\boldsymbol{n})$ is downsampled via a Transition Module $\mathcal{T}_k(\cdot)$ to generate $T_k(\boldsymbol{n})$ as the input for the next stage.
		\item Deep Supervision Branch: $\hat{X}_k(\boldsymbol{n})$ is upsampled via a Transpose Convolution $\mathcal{U}_k(\cdot)$ to generate $U_k(\boldsymbol{n})$ for global fusion.
	\end{itemize}
	
	\subsubsection{Global Fusion and Prediction}
	The decoder phase aggregates multi-scale features from all stages.
	The initial feature $\Phi_0(\boldsymbol{n})$ is refined by an independent attention module $\mathcal{A}_0(\cdot)$ to capture low-level texture cues.
	All hierarchical features are concatenated to construct the final dense representation $R(\boldsymbol{n})$:
	\begin{equation}
		\label{eq1}
		\begin{aligned}
			\Phi_0(\boldsymbol{n}) &= \mathcal{F}(I_0(\boldsymbol{n})), \\
			T_0(\boldsymbol{n}) &= \mathcal{T}_{\mathrm{init}}(\Phi_0(\boldsymbol{n})), \\
			\hat{X}_k(\boldsymbol{n}) &= \mathcal{A}_k(\mathcal{C}_k(T_{k-1}(\boldsymbol{n}))), \quad k=1, \dots, 4, \\
			T_k(\boldsymbol{n}) &= \mathcal{T}_k(\hat{X}_k(\boldsymbol{n})), \\
			U_k(\boldsymbol{n}) &= \mathcal{U}_k(\hat{X}_k(\boldsymbol{n})), \\
			R(\boldsymbol{n}) &= \text{Concat}(\mathcal{A}_{0}(\Phi_0(\boldsymbol{n})), U_1(\boldsymbol{n}), \dots, U_4(\boldsymbol{n})), \\
			P(\boldsymbol{n}) &= \text{Linear}(R(\boldsymbol{n})),
		\end{aligned}
	\end{equation}
	where $\mathcal{F}$, $\mathcal{C}_k$, $\mathcal{A}_k$, $\mathcal{T}_k$, and $\mathcal{U}_k$
	denote the Feature Extractor, Convolution Block, 3D Attention Module,
	Transition Module, and Transpose Convolution at stage~$k$, respectively.
	
	\subsection{Attention-Guided Dense Block (AGDB)}
	The Attention-Guided Dense Block (AGDB) serves as the core processing unit within HDAN.
	Unlike conventional blocks that perform sequential operations, our AGDB integrates feature extraction, refinement, and distribution.
	As illustrated in Fig.~\ref{fig2}, the AGDB comprises a Convolution Block for dense feature extraction and a 3D Attention Module for feature recalibration.
	Specifically, the input feature $T_{k-1}(\boldsymbol{n})$ first propagates through the Convolution Block to extract dense local features $X_k(\boldsymbol{n})$.
	Before any downsampling or upsampling occurs, $X_k(\boldsymbol{n})$ is modulated by the 3D Attention Module.
	This generates a spatially-refined feature $\hat{X}_k(\boldsymbol{n})$, which acts as a shared ``clean'' representation used by both the subsequent Transition Module and Transpose Convolution.
	
	\subsubsection{The Convolution Block}
	The Convolution Block aims to progressively refine high-level semantic representations.
	It is composed of four convolutional units operating in a residual manner.
	Each unit employs a pair of 3D convolutional layers ($1\times1\times1$ followed by $3\times3\times3$) to perform dimensional expansion and spatial aggregation.
	The dense connections facilitate gradient propagation and ensure comprehensive feature preservation for the attention mechanism.
	
	\subsubsection{The 3D Attention Module}
	The 3D Attention Module adaptively recalibrates volumetric feature responses across both channel and spatial domains, operating directly in 3D voxel space.
	
	\paragraph{Channel Attention}
	This branch selectively enhances discriminative channels. 
	Given a feature map $F$, a global average pooling operation aggregates channel-wise statistics, followed by two fully connected layers to generate channel weights $M_c$.
	The channel-reweighted feature is computed as $F_c = M_c \otimes F$.
	
	\paragraph{Spatial Attention} 
	Spatial attention emphasizes voxel dependencies.
	We apply both max- and average-pooling along the channel axis to form spatial descriptors.
	These are convolved to generate the spatial map $M_s$. The spatially refined feature is obtained as $F_s = M_s \otimes F_c$.
	
	\paragraph{Unified Representation} 
	The overall operation is formalized as:
	\begin{equation}
		F' = M_s(M_c(F)) \otimes F,
	\end{equation}
	where $\otimes$ denotes element-wise multiplication.
	This unified framework explicitly models voxel continuity, achieving structure-aware feature refinement that preserves local anatomical structures in low-contrast infant brain MRIs.
	
	\subsection{Other Network Components}
	
	\subsubsection{Feature Extractor}
	The Feature Extractor serves as the initial encoding stage, transforming multi-modal MRI inputs into compact volumetric representations.
	As illustrated in Fig.~\ref{fig2}, it follows the standard \textit{Convolution–Normalization–Activation} design with residual connections to ensure training stability.
	Given the dual input modalities (T1 and T2), the number of input channels is set to $2$.
	The output of this module, $\Phi_0(\boldsymbol{n})$, bifurcates into two paths: one feeds into the Initial Transition Module to start the hierarchical encoding ($T_0$), and the other passes through an independent 3D Attention Module to contribute to the final representation $R(\boldsymbol{n})$.
	
	\subsubsection{Transition Module}
	The Transition Module acts as an intermediate encoder stage that bridges low-level feature extraction and high-level semantic encoding.
	Each module contains two sequential 3D convolutional layers: a $1\times1\times1$ layer for channel projection and a $2\times2\times2$ layer for volumetric down-sampling.
	Through this compact design, the transition module filters redundant spatial details while transmitting discriminative volumetric representations ($T_k$) to the subsequent AGDB.
	
	\section{Experiments}
	
	\subsection{Datasets}
	To comprehensively evaluate the segmentation performance and clinical applicability of the proposed HDAN, we utilized two distinct datasets: the iSeg-2019 challenge dataset for model benchmarking and the Pediatric Brain MRI dataset for neurodevelopmental assessment.
	
	\subsubsection{iSeg-2019 Dataset}
	The iSeg-2019 dataset~\cite{iSeg2019} was employed to benchmark the proposed model against state-of-the-art methods. 
	The dataset consists of 10 subjects (Subjects 1--10) designated for training and 13 subjects (Subjects 11--23) designated for testing.
	Each subject includes co-registered T1- and T2-weighted MRI scans with an isotropic resolution of $1 \times 1 \times 1~\text{mm}^3$.
	The segmentation targets include white matter (WM), gray matter (GM), and cerebrospinal fluid (CSF), with ground-truth labels annotated by expert neuroradiologists.
	
	In this study, we followed the standard partition provided in the dataset distribution: Subjects 1--10 were utilized for model training, and Subjects 11--23 were used for testing.
	It is important to note that while the ground-truth labels for the testing subjects were withheld during the original challenge phase, they have since been made available in post-challenge open-source repositories for offline research evaluation.
	Therefore, all quantitative metrics reported in this paper were computed locally using these ground-truth labels, ensuring a rigorous and reproducible assessment consistent with the dataset's file structure.
	
	\subsubsection{Pediatric Brain MRI Dataset}
	The Pediatric Brain MRI~\cite{PBMRI} dataset consists of $833$ infant brain MRI scans acquired using a Siemens Magnetom Skyra 3-T MRI scanner.
	In this study, to investigate neurodevelopmental patterns, we selected a subset of $54$ samples, comprising $27$ preterm infants and $27$ full-term infants.
	This dataset serves as the basis for the clinical volumetric analysis presented in the subsequent discussion.
	
	\subsection{Evaluation Metrics}
    We quantitatively evaluate the segmentation performance using two standard metrics:
    the Dice Similarity Coefficient (Dice score) and the Modified Hausdorff Distance (MHD).
    The public evaluation tool~\cite{2015metrics} provided by the challenge organizers was
    used for computation. The Dice score measures the volumetric overlap between the
    predicted segmentation $A$ and the ground truth $G$:
    
	\begin{equation}
		\label{eq_DICE}
		\mathrm{DICE}=\frac{2\,|A \cap G|}{|A| + |G|},
	\end{equation}
	where $|A|$ and $|G|$ denote the number of foreground voxels in the prediction and ground truth, respectively.
	The DICE score ranges from $0$ to $1$, with a higher value indicating superior segmentation accuracy.
	The MHD quantifies the shape similarity by measuring the average boundary discrepancy:
	\begin{equation}
		\label{eq_MHD}
		\begin{aligned}
			\mathrm{MHD}(A,G) = \max \Bigg( \frac{1}{|A|}\sum_{a\in A}\min_{g\in G}\|a-g\|,\frac{1}{|G|}\sum_{g\in G}\min_{a\in A}\|g-a\| \Bigg),
		\end{aligned}
	\end{equation}
	where $A$ and $G$ here denote the set of points on the boundary of the predicted and ground-truth regions, respectively.
	A smaller MHD indicates smoother and more accurate boundary delineation.
	
	\subsection{Class Balancing Strategy}
	\label{sec:loss_function}
	In infant brain segmentation, the number of voxels varies significantly across tissue categories (e.g., CSF vs. Background), leading to a class imbalance problem that may cause the model to bias towards the majority class.
	To mitigate this, we employ a Weighted Cross-Entropy Loss. 
	Instead of treating all classes equally, we assign a specific weight $w_c$ to each tissue category $c$, inversely proportional to its frequency in the training set.
	Formally, the weighted loss function is defined as:
	\begin{equation}
		\label{eq_weighted_loss}
		\mathcal{L}_{\mathrm{total}} = -\frac{1}{N} \sum_{i=1}^N \sum_{c=1}^C w_c \cdot \mathbb{I}(y^{(i)}=c) \cdot \ln P_c(\boldsymbol{n}_i),
	\end{equation}
	where $N$ is the total number of voxels, $C$ is the number of classes, $y^{(i)}$ is the ground truth label, and $P_c(\boldsymbol{n}_i)$ is the predicted probability for class $c$.
	The weight $w_c$ ensures that smaller structures (such as the detailed cortex in preterm infants) contribute sufficiently to the gradient updates, thereby improving the sensitivity of the model to fine-grained anatomical details.
	
	\subsection{Implementation Details}
	As is common in most medical imaging applications, the dataset available for model training is typically limited, whereas deep learning models generally require a large amount of data~\cite{2012AlexNet,2010understanding}.
	Moreover, the proposed model usually processes full images to maintain a broad receptive field, which further exacerbates this challenge.
	To address these constraints and ensure reliable model training, we adopt a trade-off between receptive field size and dataset availability.
	Specifically, overlapping patches of size $64\times 64 \times 64$ are extracted from both the original and manually segmented images.
	To increase the number of training samples, patches are slid across the entire volume
	with a defined step size, thereby generating a sufficient number of training samples.
	Network weights are initialized with PyTorch default initialization, which is equivalent to Kaiming uniform initialization for ReLU activations \cite{He2015Delving}.
	All bias terms are initialized to zero. A coarse linear search is then performed to select the initial learning rate and weight decay parameters, with the learning rate reduced by a factor of $10$ at fixed intervals during training.
	The network is optimized using backpropagation~\cite{2002efficient}. The implementation is based on PyTorch, a popular deep learning framework.
	All experiments are implemented on an NVIDIA GeForce RTX 4090 GPU.
	
	\subsection{Quantitative Evaluation}
	In order to evaluate the superiority of the proposed model, we select $14$ baseline methods for quantitative performance comparison.
	The baseline methods include FMRIB's Automated Segmentation Tool (FAST)~\cite{2002FAST}, Majority Voting (MV)~\cite{2008MV}, Random Forest (RF)~\cite{2001RF}, RF with Auto-Context Model (LINKS)~\cite{2015LINKS}, DeepMedic~\cite{2017efficient}, 3D-UNet~\cite{20163DUNet}, CC-2D-FCN~\cite{2015FCN}, DenseVoxNet~\cite{2017DenseVoxNet}, VoxResNet~\cite{2018VoxResNet}, CC-3D-FCN~\cite{20183DSeg}, HyperDenseNet~\cite{2018HyperDenseNet}, 3D-SkipDenseSeg~\cite{2019skipconnect}, Swin-UNETR~\cite{Hatamizadeh2022SwinUNETR}, and nnU-Net~\cite{Isensee2021}.
	
	\begin{table}[h]
		\centering
		\caption{Segmentation performance comparison for Dice score between baseline methods and ours on the iSeg-2019 dataset. A higher value is better. The best performance is highlighted in bold.}
		\label{tab1_DSC}
		{
			\setlength{\tabcolsep}{3.5pt} 
			\small 
			\begin{tabular}{lccc}
				\toprule
				& White Matter & Gray Matter & Cerebrospinal Fluid\\
				\midrule
				FAST~\cite{2002FAST}             &$0.453$& $0.393$&$0.557$\\
				MV~\cite{2008MV}                 &$0.562$& $0.710$&$0.713$\\
				RF~\cite{2001RF}                 &$0.821$& $0.839$&$0.859$\\
				LINKS~\cite{2015LINKS}           &$0.839$& $0.857$&$0.885$\\
				DeepMedic~\cite{2017efficient}   &$0.844$& $0.859$&$0.917$\\
				3D-UNet~\cite{20163DUNet}        &$0.836$& $0.862$&$0.897$\\
				CC-2D-FCN~\cite{2015FCN}         &$0.756$& $0.814$&$0.854$\\
				DenseVoxNet~\cite{2017DenseVoxNet}&$0.854$& $0.879$&$0.909$\\
				VoxResNet~\cite{2018VoxResNet}   &$0.889$& $0.895$&$0.917$\\
				CC-3D-FCN~\cite{20183DSeg}       &$0.919$& $0.903$&$0.934$\\
				HyperDenseNet~\cite{2018HyperDenseNet} &$0.889$& $0.889$&$0.926$\\
				3D-SkipDensSeg~\cite{2019skipconnect}&$0.920$&$0.901$&$0.934$ \\
				Swin-UNETR~\cite{Hatamizadeh2022SwinUNETR}&$0.851$& $0.882$&$0.922$ \\
				nnU-Net~\cite{Isensee2021}       &$0.928$& $0.911$&$\boldsymbol{0.956}$ \\
				Ours  &$\boldsymbol{0.947}^*$& $\boldsymbol{0.937}^*$&$0.953$\\
				\bottomrule
				\multicolumn{4}{l}{\scriptsize{$^*$ indicates statistical significance ($p < 0.05$) compared to the runner-up method.}}
			\end{tabular}
		}
	\end{table}
	Table~\ref{tab1_DSC} reports the average similarity measured by the Dice score defined in Eq.~\eqref{eq_DICE}.
	on the testing partition (Subjects~11--23) of the iSeg-2019 dataset.
	From Table~\ref{tab1_DSC}, it is clear that the proposed HDAN outperforms the $12$ baseline methods
	in terms of white matter, gray matter, and cerebrospinal fluid.
	The proposed method achieves average Dice scores of $0.947$, $0.937$, and $0.953$ for
	white matter, gray matter, and cerebrospinal fluid, respectively.
	
	Moreover, Table~\ref{tab2_MHD} reports the average MHD defined in Eq.~\eqref{eq_MHD}
	on the iSeg-2019 dataset.
	From Table~\ref{tab2_MHD}, it clearly observed that the proposed model achieves
	the best segmentation performance in terms of the MHD metric.
	
	Comparison with SOTA: As shown in Table~\ref{tab1_DSC}, our method outperforms all CNN-based baselines.
	Furthermore, we compared our model with the self-configuring nnU-Net, a strong baseline in medical image segmentation.
	While nnU-Net achieves a slightly higher Dice score on the relatively distinct Cerebrospinal Fluid (CSF) region (0.956 vs. 0.953), our proposed method demonstrates superior performance on the most challenging tissues: White Matter (0.947 vs. 0.928) and Gray Matter (0.937 vs. 0.911).
	This significant improvement (approx. 2\% on WM) validates that our proposed attention mechanisms effectively address the low-contrast ``isointense'' problem that standard U-Net architectures struggle with.
	
	\begin{table}[h]
		\centering
		\caption{Segmentation performance comparison for the MHD between baseline methods and ours on the iSeg-2019 dataset. The lower value is better. The best performance is highlighted in bold.}
		\label{tab2_MHD}
		{
			\setlength{\tabcolsep}{3.5pt} 
			\small 
			\begin{tabular}{lccc}
				\toprule
				& White Matter & Gray Matter & Cerebrospinal Fluid\\
				\midrule
				FAST~\cite{2002FAST}             &$1.714$& $1.012$&$1.838$\\
				MV~\cite{2008MV}                 &$1.771$& $1.196$&$1.963$\\
				RF~\cite{2001RF}                 &$0.695$& $0.594$&$0.475$\\
				LINKS~\cite{2015LINKS}           &$0.569$& $0.489$&$0.439$\\
				DeepMedic~\cite{2017efficient}   &$0.519$& $0.499$&$0.475$\\
				3D-UNet~\cite{20163DUNet}        &$0.602$& $0.463$&$0.483$\\
				CC-2D-FCN~\cite{2015FCN}         &$0.789$& $0.672$&$0.567$\\
				DenseVoxNet~\cite{2017DenseVoxNet}&$0.582$& $0.456$&$0.497$\\
				VoxResNet~\cite{2018VoxResNet}   &$0.546$& $0.483$&$0.428$\\
				CC-3D-FCN~\cite{20183DSeg}       &$0.259$& $0.179$&$0.285$\\
				HyperDenseNet~\cite{2018HyperDenseNet} &$0.335$& $0.289$&$0.314$\\
				3D-SkipDensSeg~\cite{2019skipconnect}&$0.209$& $0.174$&$0.256$ \\
				Swin-UNETR~\cite{Hatamizadeh2022SwinUNETR}&$0.578$& $0.494$&$0.562$ \\
				nnU-Net~\cite{Isensee2021}       &$0.178$& $0.148$&$0.194$ \\
				Ours &$\boldsymbol{0.129}$& $\boldsymbol{0.102}$&$\boldsymbol{0.131}$\\
				\bottomrule
			\end{tabular}
		}
	\end{table}
	
	\subsection{Qualitative Evaluation}
	To provide a visual comparison of tissue segmentation performance, three representative slices were randomly selected from a sample and segmented using the proposed method, CC-3D-FCN~\cite{20183DSeg}, and 3D-SkipDenseSeg~\cite{2019skipconnect}.
	The corresponding T1-weighted MRI images and ground truth labels are also presented for reference.
	As illustrated in Fig.~\ref{fig_visualization}, the regions with notable differences are highlighted by red rectangles.
	The segmentation produced by the proposed HDAN exhibits a higher consistency with the ground truth, particularly in the delineation of white matter, compared with the two baseline approaches.
	
	\subsection{Ablation Analysis}
	To validate the contribution of individual components in HDAN, we performed a stepwise ablation study on the iSeg-2019 dataset, with quantitative results summarized in Table~\ref{tab_ablation}.
	\begin{table}[h]
		\setlength{\tabcolsep}{5.5pt}
		\caption{Ablation study of key components on the iSeg-2019 dataset. 
			``Dense Up.'' denotes Attention-Guided Dense Upsampling, ``CA'' denotes Channel Attention, and ``SA'' denotes Spatial Attention. 
			The evaluation metrics include the average Dice score ($\uparrow$) and MHD ($\downarrow$) across all tissue classes.}
		\label{tab_ablation}
		\centering
		\begin{tabular}{lccccc}
			\toprule
			\multirow{2}{*}{Method} & \multirow{2}{*}{Dense Up.} & \multirow{2}{*}{CA} & \multirow{2}{*}{SA} & \multicolumn{2}{c}{Average Metric} \\
			\cmidrule(lr){5-6}
			& & & & Dice score $\uparrow$ & MHD $\downarrow$ \\
			\midrule
			Baseline (Backbone) & \ding{55} & \ding{55} & \ding{55} & 0.865 & 0.512 \\
			+ Dense Up. Only    & \ding{51} & \ding{55} & \ding{55} & 0.872 & 0.481 \\
			+ Dense Up. + SA    & \ding{51} & \ding{55} & \ding{51} & 0.885 & 0.315 \\
			+ Dense Up. + CA    & \ding{51} & \ding{51} & \ding{55} & 0.927 & 0.183 \\
			Proposed (All) & \ding{51} & \ding{51} & \ding{51} & \textbf{0.946} & \textbf{0.121} \\
			\bottomrule
		\end{tabular}
	\end{table}
	\subsubsection{Efficacy of the Backbone Network}
	We established a baseline model (denoted as ``Baseline (Backbone)'') by removing the attention mechanisms and the dense upsampling strategy, retaining only the standard residual encoder-decoder structure.
	As presented in Table~\ref{tab_ablation}, this baseline achieved a Dice score of 0.865 and an MHD of 0.512.
	These metrics indicate that while the deep residual backbone can capture the global semantic context of brain tissues, its ability to delineate precise boundaries is compromised.
	This limitation is intrinsic to standard convolutions when applied to the isointense phase of preterm infants, where the lack of explicit feature recalibration prevents the network from resolving the subtle intensity ambiguity at tissue interfaces.
	\subsubsection{Impact of Dense Upsampling}
	Incorporating the ``Dense Up.'' strategy alone (``+ Dense Up. Only'') yielded a moderate performance gain, improving the Dice score to 0.872 and reducing the MHD to 0.481.
	This result suggests that while dense connections facilitate gradient flow and feature reuse—thereby mitigating information loss during downsampling—they are insufficient to fully resolve boundary ambiguities.
	Without an attention mechanism to filter features, dense connections inevitably propagate both task-relevant semantics and redundant background noise, thereby limiting the upper bound of segmentation accuracy.
	
	\subsubsection{Critical Role of Attention Mechanisms}
	The integration of 3D attention modules proved to be the decisive factor in performance improvement.
	Introducing Spatial Attention (SA) refined the local structural details, increasing the
	Dice score to 0.885 while moderately reducing the MHD to 0.315.
	Furthermore, the addition of Channel Attention (CA) provided a substantial boost, reaching a Dice score of 0.927 and significantly lowering the MHD to 0.183.
	This dramatic reduction in boundary error demonstrates that explicit feature recalibration—which reweights informative channels and suppresses task-irrelevant responses—is essential for distinguishing isointense tissues from the complex background.
	
	\subsubsection{Synergistic Integration}
	The fully integrated ``Proposed (All)'' configuration achieved state-of-the-art performance, with a Dice score of \textbf{0.946} and an MHD of \textbf{0.121}.
	This confirms the synergistic effect between dense upsampling and attention mechanisms: the former ensures a robust flow of multi-scale information, while the latter dynamically filters this information to focus on anatomical validity.
	This combination enables HDAN to generate highly accurate segmentation maps even in the challenging isointense phase.
	
	\subsubsection{Visual Interpretability of Attention Mechanism}
\begin{figure*}[!h]
		\centering
		\includegraphics[width=0.95\textwidth]{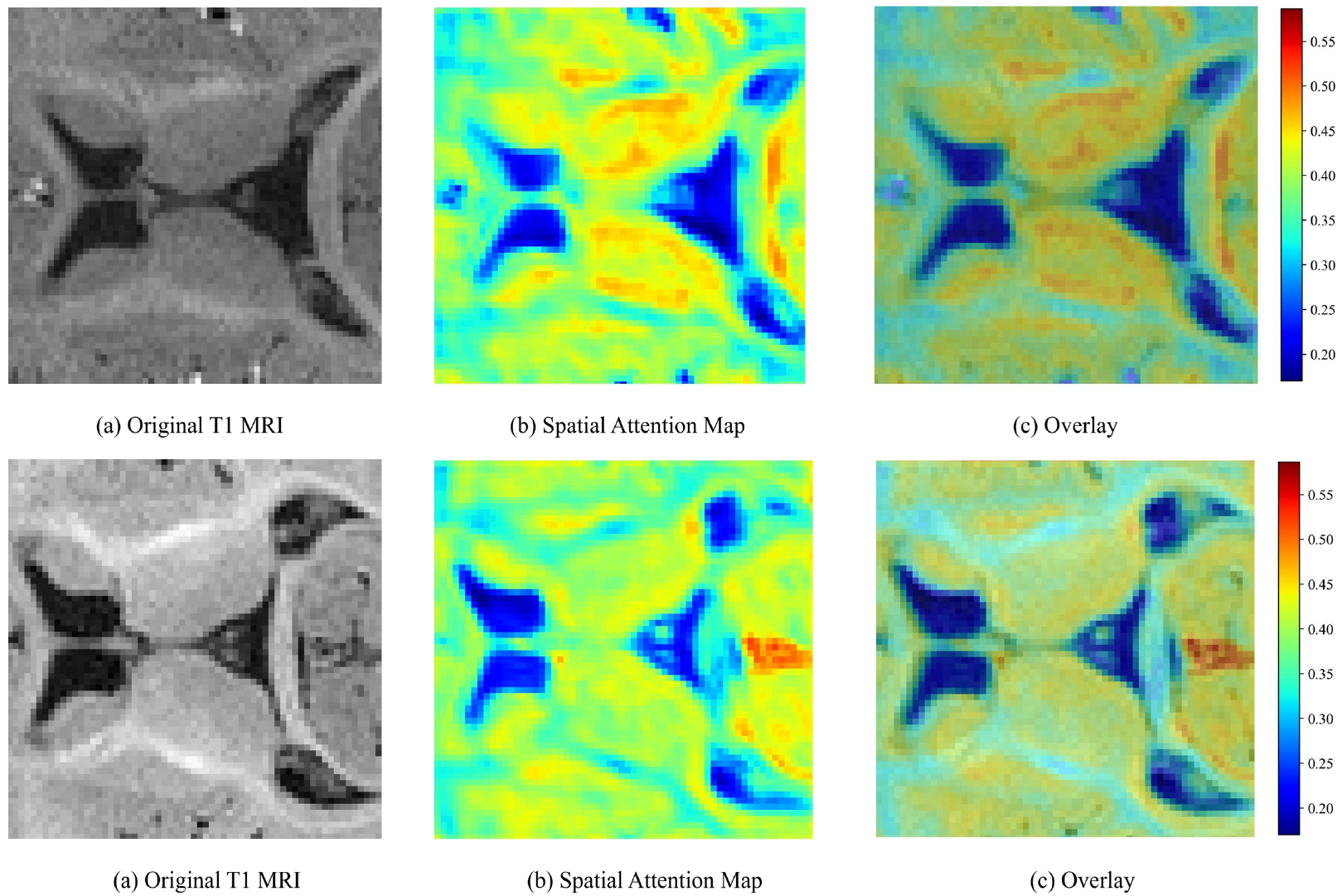} 
		\caption{Visualization of the Spatial Attention Map learned by the proposed HDAN.
			(a) The original T1-weighted MRI slice showing the low tissue contrast in the isointense phase.
			(b) The intermediate feature map extracted from the Spatial Attention module using the Jet colormap (Red: High attention; Blue: Low attention).
			(c) The overlay of the attention map on the original MRI.
			It can be clearly observed that the attention mechanism effectively suppresses the CSF within the ventricles (deep blue regions) while strongly activating along the tissue boundaries (indicated by red and yellow hues), demonstrating its capability to capture fine-grained structural details.}
		\label{fig:attention_vis}
	\end{figure*}

\begin{figure*}[!h]
    	\centering
    	\includegraphics[width=1.0\textwidth]{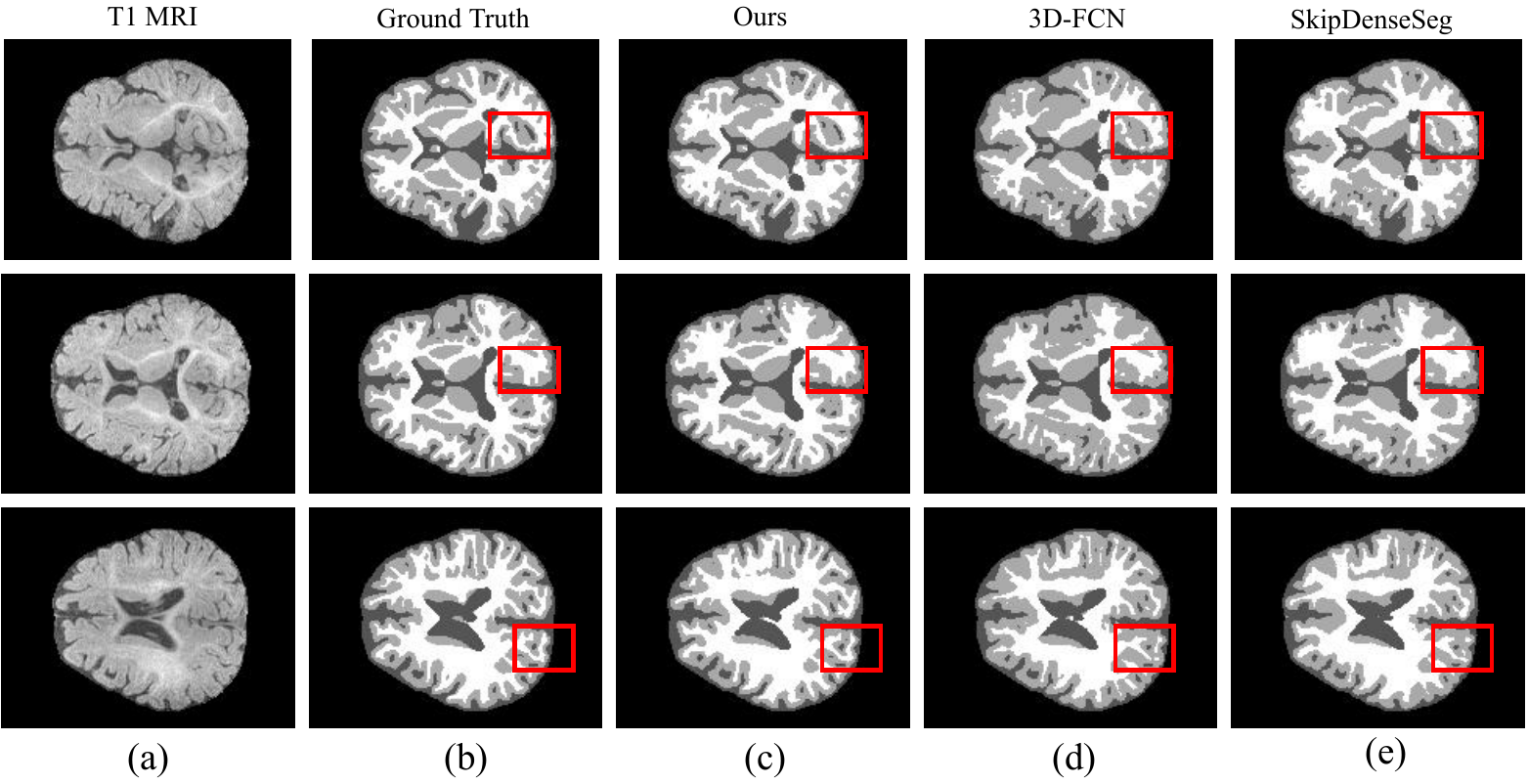}
    	\caption{Segmentation results produced by the proposed method and two baseline methods on different slices.
    		(a) T1-weighted MRI; (b) ground truth label; (c) segmentation of the proposed method;
    		(d) segmentation of 3D-FCN; (e) segmentation of SkipDenseSeg.}
    	\label{fig_visualization}
    \end{figure*}

	To elucidate why the proposed method significantly outperforms the baseline in boundary delineation (evidenced by the sharp decrease in MHD), we visualized the intermediate feature maps generated by the Spatial Attention module.
	As illustrated in Fig.~\ref{fig:attention_vis}, the attention mechanism functions as a robust feature selector.
	The visualization reveals that the module strongly suppresses the ventricular regions (Cerebrospinal Fluid, as shown in deep blue), effectively filtering out background noise.
	Conversely, the high-activation regions (highlighted in red and yellow hues) precisely track the geometry of the cortical ribbon.
	This empirical evidence corroborates that the HDAN does not merely memorize intensity values but explicitly learns the structural topology of the brain, thereby resolving the boundary ambiguity that hinders the baseline model.
	
	\subsection{Neurodevelopment Assessment in Preterm Infants}
	The primary goal of precise infant brain tissue segmentation is to enable reliable quantification of early neurodevelopment, particularly in preterm infants.
	White matter, a major component of the central nervous system, consists primarily of neuronal axons and their surrounding myelin sheaths.
	At birth, infant white matter remains immature, with incomplete myelination of axons.
	The first two postnatal years therefore represent a critical period of rapid white matter maturation~\cite{Huppi1998}.
	Clinical investigations~\cite{Gilmore2008,Ball2018} have frequently used white matter volume as a significant indicator of development progression.
	Accordingly, this study develops an efficient network-based automated framework to assess early brain development by quantifying both absolute and relative white matter volumes derived from voxel-level segmentation.
	
	Recent neuroimaging evidence~\cite{Makropoulos2018, OMuircheartaigh2020} indicates that preterm infants generally exhibit smaller total brain volume and delayed cortical (gray matter) growth compared with term-born infants.
	To examine these developmental differences, brain MRI scans of 27 preterm and 27 age-matched term infants were retrospectively collected from the Pediatric Brain MRI dataset~\cite{PBMRI}.
	All scans were preprocessed under standardized clinical protocols to ensure consistent image quality across subjects.
	The proposed segmentation model was applied to delineate brain tissues, with particular emphasis on accurate white matter segmentation.
	Since ground truth masks were unavailable for this external dataset, we performed a rigorous visual inspection of the segmentation results on randomly selected samples.
	The proposed model demonstrated robust generalization capabilities, producing anatomically plausible tissue boundaries consistent with expert visual assessment, despite potential domain shifts between datasets.
    For each subject, the voxel counts of white matter, gray matter, and cerebrospinal fluid were computed, from which both absolute and relative white matter measures
    (the white matter ratio) were derived.
    
	The averaged results are summarized in Table~\ref{tab_neural_assessment}. As presented in Table~\ref{tab_neural_assessment}, the quantitative assessment reveals significant developmental differences. The mean white matter volume of preterm infants ($649,152 \pm 42,103$ $\text{mm}^3$) was found to be significantly lower than that of term-born infants ($672,657 \pm 38,921$ $\text{mm}^3$) with $p < 0.05$.
    This volumetric reduction is consistent with the incomplete myelination status expected in preterm development~\cite{Makropoulos2018}.
    Moreover, the gray matter and total brain volumes (defined as the sum of WM and GM) were also markedly smaller ($p < 0.01$ and $p < 0.05$, respectively). Specifically, the total brain volume in the preterm group was $1,344,275$ $\text{mm}^3$, compared to $1,415,334$ $\text{mm}^3$ in the term group.
    Standardizing these metrics to physical volumes ($\text{mm}^3$) allows for direct comparison with clinical literature, confirming that our deep learning-based segmentation can reliably capture subtle volumetric deficits associated with prematurity.
    
    Such trends are consistent with established neurodevelopmental patterns—indicating delayed cortical maturation and reduced overall brain growth in preterm infants rather than tissue-specific disproportion~\cite{OMuircheartaigh2020,Huppi1998}.
	
    \begin{table}[h]
    	\centering
    	\setlength{\tabcolsep}{4.5pt}
    	\caption{Comparison of mean brain tissue volumes (mm$^3$) and white matter ratios (\%)
    		between preterm ($N=18$) and term ($N=27$) infants.}
    	\label{tab_neural_assessment}
    	\begin{tabular}{cccccc}
    		\toprule
    		& WM & GM & CSF & Brain volume\footnotemark[1] & WM ratio\footnotemark[2] \\
    		\midrule
    		Preterm & 649{,}152 & 695{,}123 & 474{,}353 & 1{,}344{,}275 & 48.15\% \\
    		Term    & 672{,}657 & 742{,}677 & 425{,}307 & 1{,}415{,}334 & 47.42\% \\
    		\textit{p}-value & $< 0.05$ & $< 0.01$ & $< 0.05$ & $< 0.05$ & $> 0.05$ \\
    		\bottomrule
    	\end{tabular}
    	\footnotetext[1]{Defined as WM $+$ GM (excluding CSF).}
    	\footnotetext[2]{\textbf{Calculated as the mean of individual subject ratios, not the ratio of the means.}}
    \end{table}
	
	\section{Conclusion}
	In this study, we proposed HDAN, a novel automatic 3D volumetric brain MRI segmentation framework for neurodevelopmental assessment in preterm infants.
	Unlike previous approaches constrained to 2D processing, our framework leverages a fully 3D convolutional architecture capable of handling multimodal MRI data.
	The standard Convolution-BatchNorm-ReLU design ensures training stability, while 3D skip connections enhance representational consistency.
	Furthermore, an integrated 3D attention mechanism improves feature discrimination and segmentation precision.
	Comprehensive quantitative and qualitative evaluations demonstrate the superior performance of the proposed model.
	When applied to preterm infant brain MRIs, the framework effectively identified characteristic developmental patterns—specifically, reduced total brain and gray matter volumes with a relatively higher white matter proportion—consistent with previously reported indicators of delayed neurodevelopment in preterm populations~\cite{Makropoulos2018,OMuircheartaigh2020,Huppi1998}.
	
	\backmatter
	
    \section*{Declarations}
    
    \bmhead{Funding}
    This work was supported by the Introduced Team Project (No. szgmtd2025001) at Shenzhen
    University of Advanced Technology General Hospital, featuring Professor Zhao Zhengyan's
    Pediatric Team from The Children's Hospital, Zhejiang University School of Medicine.
    Additional support was provided by the project titled ``AI-based assessment of brain
    maturity and brain injury in preterm infants'' (Grant No. Shen Ke Chuang Zi [2024] 47).
    
    \bmhead{Competing interests}
    The authors declare no competing interests.
    
    \bmhead{Ethics approval}
    This study was conducted using publicly available datasets and did not involve any new
    human participants or animal experiments performed by the authors. The datasets used in
    this study include the iSeg-2019 challenge dataset and the large dataset of infancy and
    early childhood brain MRIs. The original studies that collected these datasets received
    ethical approval from their respective institutional review boards, and all data were
    fully anonymized prior to public release. Therefore, no additional ethical approval was
    required for this study.
    
    \bmhead{Informed consent}
    Informed consent was obtained from the legal guardians of all participants by the original
    data providers of the publicly available datasets.
    
    \bmhead{Author contributions}
    Lexin Ren and Jiamiao Lu contributed equally to this work. All authors read and approved
    the final manuscript.
    
    \clearpage
    
	\bibliography{sn-bibliography}
	
	\end{document}